\documentclass[conference]{IEEEtran}

\usepackage{algorithm}
\usepackage{algpseudocode}
\usepackage{float}
\usepackage{flushend}
\usepackage{graphicx}
\usepackage{lettrine}
\usepackage{multirow}
\usepackage{subfigure}

\begin{document}

\title{Sensor Deployment for Air Pollution Monitoring Using Public Transportation System}

\author{James J.Q. Yu, \textit{Student Member, IEEE} \\
       Department of Electrical and\\
       Electronic Engineering\\
       The University of Hong Kong\\
       Email: jqyu@eee.hku.hk\\
\and Victor O.K. Li, \textit{Fellow, IEEE}\\
       Department of Electrical and\\
       Electronic Engineering\\
       The University of Hong Kong\\
       Email: vli@eee.hku.hk\\
\and Albert Y.S. Lam, \textit{Member, IEEE}\\
       Department of Electrical Engineering\\
       and Computer Sciences\\
       University of California, Berkeley\\
       Email: albertlam@ieee.org\\
}

\maketitle
\pagestyle{empty}

\begin{abstract}
Air pollution monitoring is a very popular research topic and many monitoring systems have been developed. In this paper, we formulate the Bus Sensor Deployment Problem (BSDP) to select the bus routes on which sensors are deployed, and we use Chemical Reaction Optimization (CRO) to solve BSDP. CRO is a recently proposed metaheuristic designed to solve a wide range of optimization problems. Using the real world data, namely Hong Kong Island bus route data, we perform a series of simulations and the results show that CRO is capable of solving this optimization problem efficiently.

\end{abstract}

\begin{keywords}
Air pollution, public transportation, evolutionary algorithm, chemical reaction optimization.
\end{keywords}

\section{Introduction}
\lettrine[lines=2]{A}{ir} pollution has raised great concern over the past few decades due to the increasing expansion of industries. It has caused many serious problems, including climate change, loss of biodiversity, changes in hydrological systems, acid rain, and stress on the system of food production \cite{WHO2005}. It is also known that some of the chemical pollutants in the air can increase the occurrence of diseases such as lung cancer and pneumonia \cite{Poschl2005}\cite{Nihal2008}. Since hazardous gases can spread over a very large region and causing huge and irreparable damage \cite{DDLee2001}, there is a growing demand for air pollution monitoring systems.

Many air pollution monitoring systems utilizing smart sensor networks and wireless systems have been proposed in the recent literature \cite{YJJung2008}\cite{MGao2008}. But most of these systems use individually designed facilities to collect and transmit detected data, and the facilities are installed on stationary bases. These systems require a large number of sensors in order to provide a satisfactory coverage of the whole area, rendering them very expensive. To increase the coverage with a limited number of sensors, one can have embedded sensors installed on moving objects, e.g., a vehicle or an animal. With the movement of these objects, a large area can be covered. Since the air pollution condition changes relatively slow and can be regarded as constant in a short period of time, the mobile sensor system can achieve a larger coverage without losing too much accuracy. Most previous work focuses on implementation aspects, e.g. the design of mobile sensor and installation on the bus \cite{FCatineira2008}\cite{FPena2010}. However, no literature has been published concerning the selection of bus routes on which sensors are to be deployed. In this paper, we propose a novel optimization problem solving this selection problem.

To select the bus routes on which sensors are to be deployed so as to minimize the number of sensors required for a satisfactory coverage, we formulate an optimization problem called Bus Sensor Deployment Problem (BSDP). To solve this problem, we employ the Chemical Reaction Optimization (CRO) technique. CRO is a population-based general-purpose optimization metaheuristic which mimics the transition and interaction of molecules in a chemical reaction. In chemical reaction there is a natural tendency for the potential energy of the reactant molecules to decrease until it reaches a stable energy state \cite{AYSLam2012}. CRO utilizes this tendency to guide molecules to explore the solution space and to find the global minimum.

In our model, we assume that the whole area is divided into square grids of the same size and we assume that the air pollution condition in the same grid are similar. The route of a bus are divided into segments according to the boundary of the grids and we consider the mid-point of these segments as the sensing point. When the bus completes its route, the stored data are uploaded wirelessly to the base station at the bus terminus.

The rest of this paper is organized as follows. Section II reviews the related work on air pollution monitoring system. The problem to be solved using CRO is described in Section III, followed by a detailed framework and algorithm design in Section IV. The simulation results are reported and discussed in Section V. Finally we conclude the paper in Section VI.

\section{Related Work}
Air pollution monitoring is a hot research topic due to the increasing concern on the adverse effects of pollution. Kularatna \textit{et al.} proposed an environmental air pollution monitoring system in \cite{Nihal2008} focusing on CO, NO$_{2}$, and SO$_{2}$ detection. The proposed system is based on a smart sensor converter installed with an application processor which can download the pollution condition for further processing. Tsow \textit{et al.} proposed a wearable and wireless sensor system for real-time monitoring of toxic environmental volatile organic compounds in \cite{Tsow2009}. Jung \textit{et al.} proposed an air pollution geo-sensor network to monitor several air pollutants in \cite{YJJung2008}. The system consists of 24 sensors and 10 routers, and provides alarm messages depending on the detected pollutants. Gao \textit{et al.} proposed a wireless mesh network to cover a given geographic area using embedded microprocessors consisting of sensors and wireless communication in \cite{MGao2008}. Kwon \textit{et al.} proposed another outdoor air pollution monitoring system in \cite{JWKwon2007}. This system uses ZigBee networks to transmit the sensed pollutant density levels. The above systems are all air pollution systems utilizing mobile sensors to achieve high coverage, but they all need proprietary equipment to accommodate the movement requirements of the system. Gil-Castineira \textit{et al.} proposed an air pollution detection system based on the public transportation system and tested it in a small scale experiment \cite{FCatineira2008}. However, there is no sensor deployment algorithm that can efficiently utilize the available resources. So in this paper, we formulate an optimization problem to deploy the sensors so as to utilize them efficiently. The optimization problem is solved with CRO.

Many optimization problems have been solved using CRO since \cite{AYSLam2010}. Xu \textit{et al.} used CRO to solve task scheduling problem in grid computing \cite{JXu2011}. This problem is a multi-objective NP-hard optimization problem. Lam \textit{et al.} proposed a population transition problem in P2P live streaming and solved this problem using CRO in \cite{AYSLam2010b}. Lam and Li also solved the cognitive radio spectrum allocation problem in \cite{AYSLam2010c}. Several variants of CRO were proposed to solve the optimization problem and a self-adaptive scheme was used to control the convergence speed of CRO \cite{AYSLam2010c}. Yu \textit{et al.} proposed a CROANN algorithm from real-coded version of CRO \cite{JJQYu2011} to train artificial neural networks (ANNs). CROANN  used a novel stopping criteria to prevent the ANNs from being over-trained and the simulation results demonstrated that CROANN outperformed most previously proposed EA-based ANNs training methods as well as some sophisticated heuristic training methods. This shows that CRO has great potential to tackle different optimization problems like BSDP discussed in this paper.

\section{Problem Formulation}
In this paper, we formulate BSDP as a grid coverage problem. The monitoring area $A$ is divided into $p\times q$ grids. There are $n$ bus routes and each bus route $R_{i}, i=1,2,...,n$ passes through multiple grids. If a bus route passes through a grid, then we say this bus route ``covers" the grid. We then define a coverage threshold $c$ such that a grid is ``fully-covered" if it can be covered by $c$ or more routes. A solution $S$ is given by a vector $S=[s_{1}, s_{2}, ..., s_{n}]$, where $s_{i}=1$ means there is a sensor installed on the buses for $R_{i}$ and $s_{i}=0$ means no sensors installed for $R_{i}$. With all the bus routes' sensor-installation information $s_{i}$ collected, we can make a coverage graph $G_{S_{f}}$ where the solution $S_{F}=[1, 1, ..., 1]$ means all bus routes are equipped with sensors, and we define the total number of covered grids in $G_{S_{f}}$ is $t_{S_{f}}$. Then given a random solution $S^{\prime}$ we can compute its coverage graph $G_{S^{\prime}}$ and its covered grid number $t_{S^{\prime}}$. It is natural to assume that the pollutant level in each grid remains constant in a short period of time like one hour \cite{Nihal2008} and we set the sensing interval for each grid to be one hour. Since all buses we use for simulation can finish their routes in this period of time, we consider that there is only one bus with sensor installed running on the same route simultaneously, and the bus can pass all the grids in each sensing interval. Take Fig. 1 as an example. The two black circles are two bus routes. The 28 grids with shading are covered by at least one route. The four grids in the center with darker shading are covered by two routes. So in this example if we set $c=$ 1 or 2, $t_{S_{f}}=$ 28 or 4, respectively. In later calculation we only consider these $t_{S_{f}}$ grids. If we only install sensors on buses running on the left circle, the $t_{S^{\prime}}$ is 16 or 4 with $c=$ 1 or 2. If we desire to have a shorter sensing interval, say $m$ times instead of once per hour, we can deploy sensors on $m$ of the buses on each route, with bus start times of $1/m$ hours apart.

\begin{figure}
	\centering
	\includegraphics[width=0.5\textwidth]{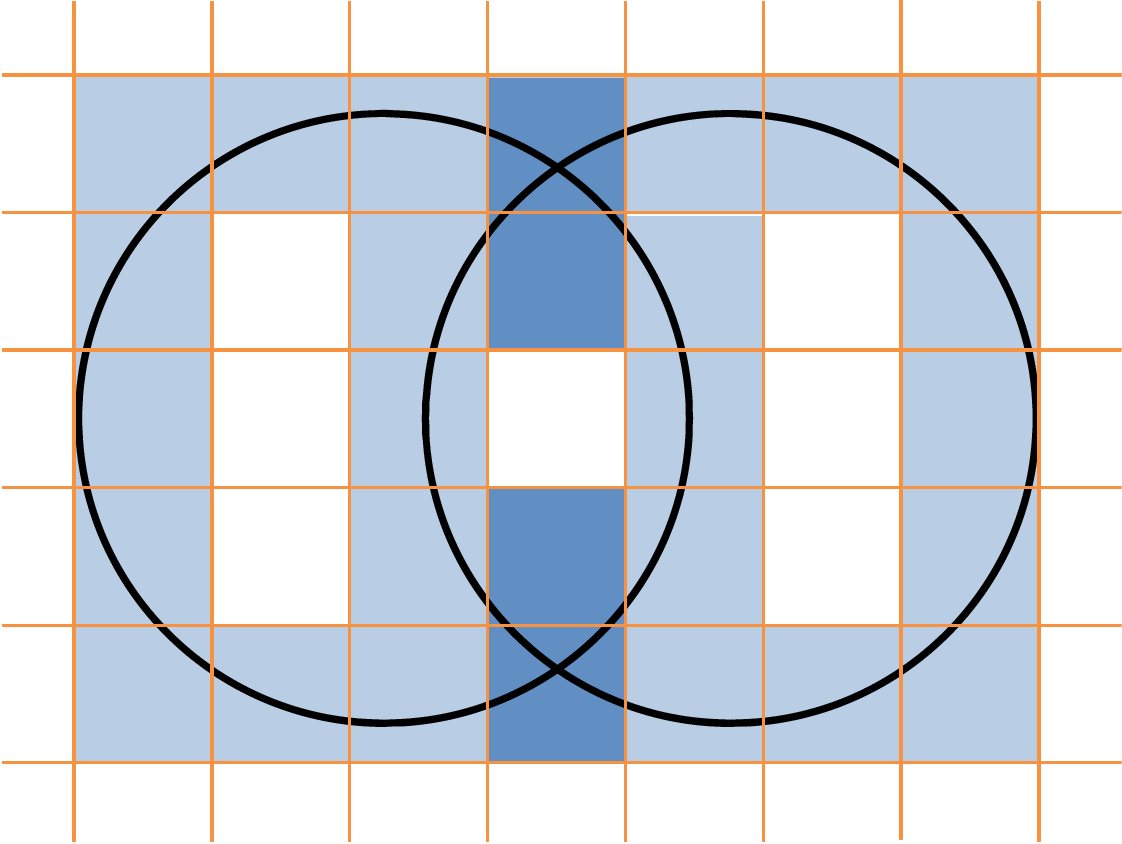}
	\caption{An Example of Problem Instance}
\end{figure}

In this problem we have two major factors to evaluate a solution: the coverage percentage and the total sensor number. The former concerns the performance as well as data accuracy of the system while the latter concerns the expense of deploying the system. Since in different scenarios the system may have different requirements, we introduce a weight coefficient $\alpha$ to balance the two factors. The objective function for BSDP is given as follows:

\begin{equation}
\min\hspace{3mm}(1-\frac{t_{S^{\prime}}}{t_{S_{f}}})\times\alpha + \frac{\sum_{i=1}^{n}s_{i}}{n}\times (1-\alpha).
\end{equation}

This objective function is composed of two parts: the percentage of uncovered grids over the total coverable grids $\frac{t_{S^{\prime}}}{t_{S_{f}}}$ and the percentage of sensor deployment $\frac{\sum_{i=1}^{n}s_{i}}{n}$. In this paper, we adopt $\alpha=0.5$ in our simulations in Section V to simulate a general case that both the coverage and the cost of purchasing sensors are important.

\section{Algorithm Design}
In this section, the detailed design of our algorithm to solve BSDP is given. First we briefly introduce how CRO works, and then the encoding scheme and operators employed for optimization are presented.

\subsection{Chemical Reaction Optimization}
CRO is a kind of variable-population-size-based meta-heuristics that can solve optimization problems efficiently. CRO exploits the natural tendency of chemical reactions to reduce the total potential energy in reactant molecules to search the solution space and to find the global optimum. In a chemical reaction process, the molecules with potential energy (PE) as well as kinetic energy (KE) are put into a closed container. When a collision happens, the molecules will change their structure to transform PE to KE or vice versa, or or just release the energy to the environment. If we consider the energy states of molecules as a surface, this procedure can be considered as molecules rolling down the energy surface to reach the lowest energy state. CRO utilizes this natural tendency to perform optimization.

In CRO, each molecule has a molecular structure $\omega$ and two kinds of energy, i.e. KE and PE. The molecular structure stands for a feasible solution to the problem, PE is the objective function value for the solution, and KE is set as a tolerance for the molecule to move to another energy state with higher energy. We use four different types of elementary reactions to imitate all kinds of molecular collisions, namely, on-wall ineffective collision, decomposition reaction, inter-molecular ineffective collision, and synthesis reaction. The four elementary reactions cooperate with each other to search the minimums while maintaining a wide population diversity.

When CRO algorithm starts, some randomly generated molecules are initially put into a closed container. Then in each iteration one collision takes place in the container. The collision can either be one molecule colliding on the wall, or two molecules colliding with each other. We divide the four different elementary reactions according to the molecules involved in the collision into two categories: uni-molecular collisions and inter-molecular collisions, and we first randomly select one category. The former includes the on-wall ineffective collision and decomposition, while the latter includes the inter-molecular ineffective collision and synthesis. After the reaction category has been decided, the system will randomly select molecule(s) to participate in the chemical reaction. The system then check the energy of the selected molecule(s) to determine which exact elementary reaction shall happen and then the corresponding operator is performed on the involved molecule(s). The final step before the end of each iteration is the performance check. The objective function value(s) of the newly generated molecule(s) is computed and compared with previous value(s). If the new value(s) can satisfy the energy conservation conditions discussed in \cite{AYSLam2010}, the new molecule(s) is accepted and substitutes the original molecules. Otherwise the new molecule(s) is discarded. This completes an iteration of CRO. After the number of iterations reaches a certain number or other stopping criteria is met, the algorithm terminates. Interested readers can refer to \cite{AYSLam2010} and \cite{AYSLam2011} for detailed description of the algorithm as well as its pseudocode.

\subsection{Encoding Scheme}
We use a simple encoding scheme to encode a solution of BSDP. A solution is formulated as a vector of $n$ binary numbers. Each element in the solution stands for whether to install sensors for the specific route or not.

\subsection{Operators}
CRO has four different types of elementary reactions, which correspond to different functionalities. So we design a corresponding operator for each of them. We also design an initial solution generator to generate the solution structures of new molecules. We generate all random numbers uniformly in the given ranges, unless stated otherwise.

\subsubsection{Initial Solution Generator}
This initial solution generator is designed to generate new molecular structures, which is triggered when CRO is initialized or a decomposition happens \cite{AYSLam2010}. We randomly assign 0 or 1 to each element in the vector to generate new molecules. Its pseudocode is given in Algorithm 1 below:

\begin{algorithm}
\caption{\sc{InitialGen} ($\omega$)}
	\begin{algorithmic}[1]
	\ForAll{Elements $\epsilon$ in $\omega$}
		\State Randomly generate a real number $n\in [0,1)$.
		\If {$n<0.5$}
			\State $\epsilon=0$
		\Else
			\State $\epsilon=1$
		\EndIf
	\EndFor
	\end{algorithmic}
\end{algorithm}

\subsubsection{Neighborhood Search Operator}
This operator is applied to the two ineffective reactions, namely the on-wall ineffective collision and the inter-molecular ineffective collision. It is designed to generate a new molecular structure $\omega^{\prime}$ from the neighborhood of the given molecular structure $\omega$. Its main purpose is to perform a detailed local search for potentially better solutions \cite{AYSLam2010}. A random-toggle scheme is used to perform this operation. We first randomly pick an element $\epsilon_{i}$ from $\omega$ and then update the value by $\epsilon_{i}=1-\epsilon_{i}$. This operation can efficiently perform a neighborhood search on the solution space without losing accuracy. The pseudocode of this operator is given in Algorithm 2 below:

\begin{algorithm}
\caption{\sc{Ineffective} ($\omega$)}
	\begin{algorithmic}[1]
	\State Generate a random integer $i$ smaller than the total number of elements in a solution
	\State Find the $i^{th}$ element $\epsilon_{i}$ in $\omega$
	\State $\epsilon_{i}=1-\epsilon_{i}$
	\end{algorithmic}
\end{algorithm}

In an on-wall ineffective collisions, one molecule is involved and this operator can be directly applied to change the molecular structure. However, in an inter-molecular ineffective collision, since two molecules are involved, we separately manipulate them using this operator to perform local search.

\subsubsection{Decomposition}
This operator is used for generating two new molecular structures $\omega_{1}^{\prime}$ and $\omega_{2}^{\prime}$ from the given molecular structure $\omega$. This operator mainly focuses on helping the algorithm to jump out of local minimums \cite{AYSLam2010} by making severe changes with energy sharing. The pseudocode of this operator is given in Algorithm 3 on the next page.

\begin{algorithm}
\caption{\sc{Decomposition} ($\omega$)}
	\begin{algorithmic}[1]
	\State Copy $\omega$ to $\omega_{1}^{\prime}$ and $\omega_{2}^{\prime}$
	\ForAll{Elements $\epsilon$ in $\omega_{1}^{\prime}$'s and $\omega_{2}^{\prime}$'s molecular structure}
		\State Randomly generate a real number $n\in [0,1)$.
		\If {$n<0.5$}
			\State $\epsilon=1-\epsilon$
		\EndIf
	\EndFor
	\end{algorithmic}
\end{algorithm}

The original molecule is copied to two new molecules and each element in the new molecular structure is individually toggled with a probability of 0.5. If this operation can satisfy the energy balance rule mentioned in \cite{AYSLam2010} then this reaction is accepted, i.e. the original molecule is discarded and the two new molecules are put into the container.

\subsubsection{Synthesis}
This operator is used for generating one new molecular structure $\omega^{\prime}$ from two given molecular structure $\omega_{1}$ and $\omega_{2}$. This operator can perform a general local search while preventing the molecules from being stuck in the local minimums \cite{AYSLam2010}. The pseudocode of this operator is given in Algorithm 4 below:

\begin{algorithm}
	\caption{\sc{Synthesis} ($\omega_{1}$, $\omega_{2}$)}
	\begin{algorithmic}[1]
		\ForAll{Elements $\epsilon$ in $\omega^{\prime}$'s molecular structure}
			\State Randomly generate a real number $n\in [0,1)$.
			\If{$r>0.5$}
				\State $\epsilon=$counterpart in $\omega_{1}$
			\Else
				\State $\epsilon=$counterpart in $\omega_{2}$
			\EndIf
		\EndFor
	\end{algorithmic}
\end{algorithm}

In this operator, the new molecule is composed of the two given original molecules and each element of the solution is equally likely to be selected from each of the original molecules.

\begin{table*}
	\caption{Analysis on Performance for Different CRO Parameters}
	\small
	\begin{center}
		\begin{tabular}{rlll||rlll||rlll}
			\hline\hline
			\multicolumn{4}{c||}{EnBuff} & \multicolumn{4}{c||}{IniKE} & \multicolumn{4}{c}{CollRate} \\ \hline
			Value & Mean & Std. & Best & Value & Mean & Std. & Best & Value & Mean & Std. & Best \\ \hline\hline
			0 & 0.423302 & 0.0863343 & 0.398626 & 0 & 0.428764 & 0.11517 & 0.387637 & 0.1 & 0.429088 & 0.0929551 & 0.398626 \\
			10 & 0.422423 & 0.102802 & 0.395604 & 10 & 0.430577 & 0.158681 & 0.387637 & 0.2 & 0.429736 & 0.111531 & 0.387637 \\
			100 & 0.423159 & 0.0828353 & 0.395604 & 100 & 0.425835 & 0.111766 & 0.395604 & \bf{0.4} & 0.428396 & 0.0933636 & 0.401648 \\
			1000 & 0.428346 & 0.0879891 & 0.398626 & 1000 & 0.427692 & 0.112578 & 0.398626 & 0.6 & 0.43139 & 0.0969814 & 0.409615 \\
			\bf{5000} & 0.428121 & 0.0823935 & 0.40467 & \bf{5000} & 0.427286 & 0.102947 & 0.387637 & 0.8 & 0.426929 & 0.0986975 & 0.398626 \\
			10000 & 0.423099 & 0.0932018 & 0.37967 & 10000 & 0.431253 & 0.108105 & 0.398626 & 0.9 & 0.430945 & 0.108942 & 0.398626 \\ \hline\hline
			\multicolumn{4}{c||}{LossRate} & \multicolumn{4}{c||}{DecThres} & \multicolumn{4}{c}{SynThres} \\ \hline
			Value & Mean & Std. & Best & Value & Mean & Std. & Best & Value & Mean & Std. & Best \\ \hline\hline
			0.1 & 0.432907 & 0.125838 & 0.398626 & 100 & 0.450401 & 0.147143 & 0.409615 & 10 & 0.411236 & 0.0971248 & 0.376648 \\
			0.2 & 0.43133 & 0.127488 & 0.384615 & \bf{300} & 0.409709 & 0.091254 & 0.37967 & 50 & 0.413066 & 0.0835925 & 0.384615 \\
			0.4 & 0.423527 & 0.109854 & 0.387637 & 500 & 0.413154 & 0.0904357 & 0.387637 & \bf{100} & 0.406973 & 0.0803982 & 0.387637 \\
			0.6 & 0.414357 & 0.0957439 & 0.384615 & 1000 & 0.410253 & 0.0943316 & 0.376648 & 300 & 0.410703 & 0.094436 & 0.376648 \\
			\bf{0.8} & 0.410626 & 0.0908277 & 0.376648 & 3000 & 0.419896 & 0.121163 & 0.37967 & 500 & 0.41239 & 0.0972623 & 0.387637 \\
			0.9 & 0.422995 & 0.093694 & 0.387637 & 5000 & 0.427819 & 0.210808 & 0.376648 & 1000 & 0.411368 & 0.0983823 & 0.387637 \\ \hline\hline
		\end{tabular}
	\end{center}
\end{table*}

\section{Simulation Results}
In this section we will first introduce the data used for simulation of BSDP. Then the detailed simulation parameter settings, results, and comparisons are presented.

\subsection{Simulation Data}
In order to make the simulation results persuasive, we utilize the real data of the Hong Kong Island bus routes in the simulation. We selected 91 bus routes from the Citybus transportation system \cite{CityBus2011} which covers most of the accessible areas on Hong Kong Island. There are totally 2277 stations for the 91 bus routes and for simplicity, but without losing generality, adjacent stations on the same routes are directly connected and buses are supposed to run on the connected graph. The plot of all routes is given in Fig. 2.

\begin{figure}
	\centering
	\includegraphics[width=0.5\textwidth]{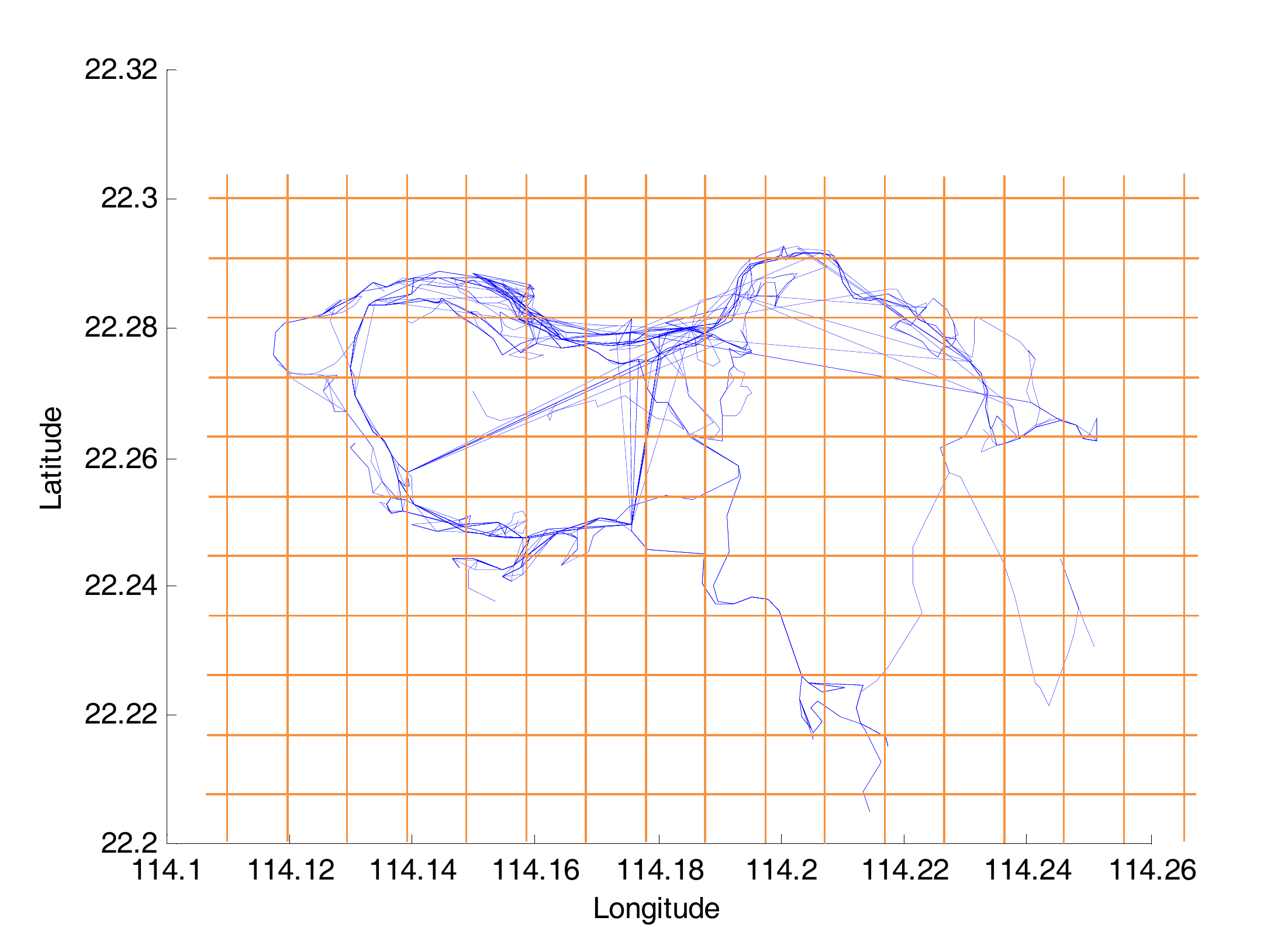}
	\caption{Grids and 91 Selected Routes for Simulation and}
\end{figure}

The whole area is divided into $16\times11=176$ $1km\times1km$ grids. For different coverage threshold $c$ we have different $t_{S_{f}}$. For instance, if $c=5$ then $t_{S_{f}}=40$. The reason why $t_{S_{f}}$ cannot reach 176 is that there are no bus routes passing through some rural grids.

\subsection{Analysis of CRO parameters selection}
The performance of CRO is greatly influenced by the proper selection of the optimization parameters \cite{JJQYu2011}. The ratio of occurrence of different reactions, the tolerance of molecules to jump to a high energy state and the energy consumption rate are all key factors that drive significant impact on the quality of the final solutions. So it is essential to analyze and select a proper combination of parameters for the simulation. In this analysis, we use the Hong Kong Island bus route data and set $c=5$. The results of testing the parameters are generated by computing the average of 50 trials with 10 000 function evaluations. Here are brief introductions to the six parameters and interested reader can refer to \cite{AYSLam2010} and \cite{AYSLam2011} for more information.

\subsubsection{EnBuff}
This parameter describes the initial energy buffer size of the container. When the algorithm iterates, this energy can be transfered from and to the molecules' PE or KE.

\subsubsection{IniKE}
This parameter describes the initial kinetic energy each molecule holds. It decides the tolerance of accepting bad molecules.

\subsubsection{CollRate}
This parameter describes the fraction of an elementary reactions being an inter-molecular collision. It functions when an iteration starts and decide the category of reactions for this iteration. The larger this parameter is, the larger possibility that inter-molecular collision will happen.

\subsubsection{LossRate}
This parameter describes the energy loss rate when an on-wall ineffective collision happens.

\subsubsection{DecThres}
This parameter describes a threshold for decomposition reaction. This parameter was named ``alpha" in \cite{AYSLam2010}. When the iteration is decided to be a uni-molecular collision, the algorithm will decide whether the decomposition reaction shall be conducted or not by using this parameter.
\subsubsection{SynThres}
Similar to DecThres, this parameter describes the happening threshold for synthesis reaction. This parameter was named ``beta" in \cite{AYSLam2010} and operates when inter-molecular collision is selected.

From the analysis results listed in Table I we can see that although the performance of the simulation can stay at a relatively high level, proper parameters can improve the final performance in a small scale. So we select a combination of parameters whose performance in mean result as well as the standard derivation is relative good. The proper combination for simulation is listed in Table II.

\begin{table}
	\caption{CRO Parameters}
	\small
	\begin{center}
		\begin{tabular}{r||l}
		\hline\hline
		Parameters & Values \\ \hline
		Function Evaluation Limit & 10 000\\ 
		Initial Population Size & 20 \\
		Initial Energy Buffer Size & 5000 \\
		Initial Molecular Kinetic Energy & 5000 \\
		Molecular Collision Rate & 0.4 \\
		Kinetic Energy Loss Rate & 0.8 \\
		Decomposition Threshold & 300 \\
		Synthesis Threshold & 100 \\ \hline\hline
		\end{tabular}
	\end{center}
\end{table}

\begin{table*}
	\caption{Analysis of the Impact of $c$ on the Results}
	\small
	\begin{center}
		\begin{tabular}{r||lll|cl|ll|lc}
			\hline\hline
			\multirow{2}{*}{$c$} & \multicolumn{3}{c|}{Objective Function Result} & \multicolumn{2}{c|}{Full-coverage Result} & \multicolumn{2}{c|}{Average Result} & \multicolumn{2}{c}{Best Result} \\ \cline{2-10}
			& Mean & Std. & Best & Covered Grids & Percentage & Coverage & Routes Count & Coverage & Routes Count \\ \hline
			1 & 0.153231 & 0.0291638 & 0.142857 & 65 & 36.9318\% & 99.7538\% & 13.72 & 100.000\% & 13 \\
			2 & 0.215108 & 0.0434328 & 0.201872 & 54 & 30.6818\% & 97.8889\% & 17.70 & 96.2963\% & 15 \\
			3 & 0.278838 & 0.0556753 &0.262166 & 49 & 27.8409\% & 97.9184\% & 23.48 & 97.9592\% & 22 \\
			4 & 0.359091 & 0.0953412 & 0.329670 & 44 & 25.0000\% & 98.0909\% & 30.94 & 100.000\% & 30 \\
			5 & 0.410451 & 0.0836293 & 0.387637 & 40 & 22.7273\% & 95.9000\% & 33.62 & 97.5000\% & 33 \\
			6 & 0.482267 & 0.1541760& 0.452574 & 38 & 21.5909\% & 91.1579\% & 35.84 & 92.1053\% & 34 \\
			\hline\hline
		\end{tabular}
	\end{center}
\end{table*}

\begin{table*}
	\caption{Comparison among CRO, SRM, and SGA}
	\small
	\begin{center}
		\begin{tabular}{r||lll|lll|lll}
			\hline\hline
			\multirow{2}{*}{$c$} & \multicolumn{3}{c|}{CRO Result} & \multicolumn{3}{c|}{SRM Result} & \multicolumn{3}{c}{SGA Result} \\ \cline{2-10}
			& Mean & Std. & Best & Mean & Std. & Best & Mean & Std. & Best \\ \hline
			1 & \textbf{0.153231} & 0.0291638 & 0.142857 & 0.401363 & 0.119488 & 0.389011 & 0.40633 & 0.150758 & 0.323077 \\
			2 & \textbf{0.215108} & 0.0434328 & 0.201872 & 0.450981 & 0.113744 & 0.455230 & 0.455551 & 0.136338 & 0.407204 \\
			3 & \textbf{0.278838} & 0.0556753 &0.262166 & 0.508006 & 0.104393 & 0.475667 & 0.522512 & 0.127871 & 0.459969 \\
			4 & \textbf{0.359091} & 0.0953412 & 0.329670 & 0.55979 & 0.095316 & 0.563437 & 0.567597 & 0.084297 & 0.53047 \\
			5 & \textbf{0.410451} & 0.0836293 & 0.387637 & 0.587005 & 0.110423 & 0.575549 & 0.599115 & 0.107698 & 0.561538 \\
			6 & \textbf{0.482267} & 0.1541760& 0.452574 & 0.668028 & 0.129847 & 0.663389 & 0.670804 & 0.132876 & 0.621747 \\
			\hline\hline
		\end{tabular}
	\end{center}
\end{table*}

\subsection{Analysis on the Impact of $c$ to Result}
To analyze the impact of $c$ on the BSDP optimization result, we adopt different values of $c$ and perform simulations using the previously stated parameter combination. The results are listed in Table III and are generated by computing the average of 50 trials with 10 000 function evaluations.

In Table III, the first column ``Objective Function Result" presents the raw data obtained from the objective function listed in Section III. The ``Full-coverage Result" column presents $t_{S_{f}}$ described in Section III, with respect to different value of $c$. If the covered grid count is high, then the largest possible cover  percentage over the whole area is also high. Since in the Hong Kong Island data there are many grids uncovered (because of no inhabitation or sea area), the percentages are relatively low. The ``Average Result" presents the mature data of the average value of the 50 best result generated from the trials. The ``Best Result" presents the mature data for the best-performing result from the 50 best result.

The selection of $c$ has a large impact on the final result. Since $c$ defines the coverage threshold, a larger $c$ will result in more buses passing through a grid. However the total number of bus routes is a fixed number, and in some remote area it is possible that there are totally less than $c$ bus routes. On the contrary, a smaller $c$ can result in a possibly higher coverage rate, the collected data accuracy is less. So the selection of $c$ shall be decided by the real-world requirement of the system.

\subsection{Comparison among Chemical Reaction Optimization, A Random Method, and Simple Genetic Algorithm}
We compare the performance of CRO with a greedy approach called Simple Random Method (SRM), and with Simple Genetic Algorithm (SGA). SRM generates 10 000 random solutions and stores the best-so-far solution. For each element in the solution SRM will perform the randomization elaborated below. First we generate a random number $n\in [0,1)$ as ``base", then we generate another random number $m\in [0,1)$ independently as ``target". If $m>n$ then this element is set to 1; otherwise, the element is set to 0. The reason why we generate two random numbers instead of one for randomization is that we cannot set a fixed threshold controlling the ratio between sensor-equipped routes and other routes. But with this double-randomization technique, the elements are individually generated and all have different probabilities of being 1. The pseudocode of SRM is given in Algorithm 5 below:

\begin{algorithm}
	\caption{\sc{SRM}}
	\begin{algorithmic}[1]
		\State Set $GlobalMin$ to be a large number
		\While{Function evaluation count is not exhausted}
			\State Initiate a new solution $s$
			\ForAll{Elements $\epsilon$ in $s$}
				\State Randomly generate a real number $n\in [0,1)$.
				\State Randomly generate another real number $m\in [0,1)$.
				\If{$m>n$}
					\State $\epsilon=1$
				\Else
					\State $\epsilon=0$
				\EndIf
			\EndFor
			\If{The objective function value of $s$ is smaller than $GlobalMin$}
				\State $GlobalMin=$ objective function value of $s$
			\EndIf
		\EndWhile
	\end{algorithmic}
\end{algorithm}

We follow \cite{CJ2012} to program SGA and we adopt the crossover rate as 0.5 and permutation rate as 0.1 for this simulation. For all three algorithms, we set the function evaluation limits to 10 000. Since the parameter $c$ shall be decided by the real-world requirement, we compare the performance of CRO, SRM and SGA with $c \in \{1,2,3,4,5,6\}$, respectively. All the simulation results are presented in Table IV.

From Table IV, we can see that CRO can outperform SRM and SGA dramatically with different values of $c$ for the mean, the standard deviation and the best result. This shows that CRO is an efficient way in solving BDSP optimization problem.

\section{Conclusion and Future Work}
In this paper, we propose a novel air pollution monitoring system by deploying sensors in a public transportation system. We also formulate a new optimization problem for selecting the buses to deploy the sensors, called BSDP. The main idea is to install sensors on buses and with the movement of the buses, the sensors can cover a much larger area compared with stationary sensor stations. This raises the problem of selecting buses to install sensors. We use CRO to solve this optimization problem since CRO has been shown to be powerful in optimizing similar problems \cite{AYSLam2010}\cite{JJQYu2011}. In our simulation, we use the Hong Kong island bus route data to analyze the impact of different parameters on the final result of BSDP optimization. CRO is also compared with SGA and a greedy method, SRM. Simulation results show that the proper selection of $c$ has great impact on the final optimization result, namely, the total number of sensors needed, the area coverage percentage, as well as the data accuracy. Meanwhile CRO can outperform  SGA and SRM with different values of $c$, in both average quality of solutions and the best generated solution.

In the future we will conduct a systematic analysis on the variance of the different parameters and perform Student's t-test on the variance. Then we will try to use metaheuristics other than CRO and SGA to see which metaheuristic is the most effective in solving BSDP. Moreover, the paper can be further extended in several ways. One possible extension is to have different number of sensors installed on different routes to reduce the sensing interval and increase the accuracy. Another is to set different coverage thresholds for different grids (regions) to reflect the different time constants for pollutant level changes. We will also deploy the sensors on a real transportation system for real-world testing.

\section*{Acknowledgement}
This work is supported in part by the Initiative on Clean Energy and Environment of The University of Hong Kong. A.Y.S. Lam is also supported in part by the Croucher Foundation Research Fellowship.


\begin{thebibliography}{99}
\bibitem{WHO2005} WHO, ``Global Environmental Change," World Health Organization, Geneva, Switzerland, 2005.

\bibitem{Poschl2005} U. Poschl, ``Atmospheric aerosols: Composition, transformation, climate and health effects," \textit{J. Atmospheric Chem. Sci.}, vol. 44, pp. 7520-7540, Nov. 2005.

\bibitem{Nihal2008} N. Kularatna and B. H. Sudantha, ``An Environmental Air Pollution Monitoring System Based on the IEEE 1451 Standard for Low Cost Requirements," \textit{IEEE Sensors J.}, vol. 8, no. 4, pp. 415-428, Apr. 2008.

\bibitem{DDLee2001} D. D. Lee and D. S. Lee, ``Environmental gas sensors," \textit{IEEE Sensors J.}, vol. 1, no. 3, pp. 214-215, Oct. 2001.

\bibitem{YJJung2008} Y. J. Jung, Y. K. Lee, D. G. Lee, K. H. Ryu, and S. Nittel, ``Air pollution monitoring system based on geosensor network," in \textit{Proc. IEEE Int. Geoscience Remote Sensing Symp.}, 2008, vol. 3, pp. 1370-1373.

\bibitem{MGao2008} M. Gao, F. Zhang, and J. Tian, ``Environmental monitoring system with wireless mesh network based on embedded system," in \textit{Proc. 5th IEEE Int. Symp. Embedded Comput.}, 2008, pp. 174-179.

\bibitem{FCatineira2008}F. Gil-Castineira, F.J. Gonzalez-Castano, R. J. Duro, and F. Lopez-Pena, ``Urban Pollution Monitoring through Opportunistic Mobile Sensor Networks Based on Public Transport," in \textit{Proc. IEEE CIMSA'08}, 2008, pp. 70-74.

\bibitem{FPena2010}F. Lopez-Pena, G. Varela, A. Paz-Lopez, R. J. Duro, and F.J. Gonzalez-Castano, ``Public Transportation Based Dynamic Urban Pollution Monitoring System," \textit{Sensors \& Transducers Journal}, vol.8, pp. 13-25, Feb. 2010.

\bibitem{AYSLam2012}A. Y.S. Lam and V. O.K. Li, ``Chemical reaction optimization: A tutorial," Memetic Computing, vol. 14, no. 1, pp. 3-17, Mar. 2012.

\bibitem{Tsow2009} F. Tsow, E Forzani, A. Rai, R. Wang, R. Tsui, S. Mastroianni, C.Knobbe, A. J. Gandolﬁ, and N. J. Tao, ``A wearable and wireless sensor system for real-time monitoring of toxic environmental volatile organic compounds,” \textit{IEEE Sensors J.}, vol. 9, pp. 1734–1740, Dec. 2009.

\bibitem{JWKwon2007} J. W. Kwon, Y. M. Park, S. J. Koo, and H. Kim, ``Design of air pollution monitoring system using ZigBee networks for ubiquitous-city,” in \textit{Proc. Int. Conf. Convergence Information Technology}, 2007, pp. 1024–1031.

\bibitem{AYSLam2010} A. Y.S. Lam and V. O.K. Li, ``Chemical-reaction-inspired metaheuristic for optimization," \textit{IEEE Trans. Evol. Comput.}, vol. 14, no. 3, pp. 381-399, Jun. 2010.

\bibitem{JXu2011} J. Xu, A. Y.S. Lam and V. O.K. Li, ``Chemical Reaction Optimization for Task Scheduling in Grid Computing," \textit{IEEE Trans. Para. Dist. Sys.}, vol.22, no. 10, pp.1624-1631, Oct. 2011.

\bibitem{AYSLam2010b} A. Y.S. Lam, J. Xu, and V. O.K. Li, ``Chemical Reaction Optimization for Population Transition in Peer-to-peer Live Streaming," in \textit{Proc. IEEE Congr. Evol. Comp.}, Jul. 2010, pp. 1-8.

\bibitem{AYSLam2010c} A. Y.S. Lam, and V. O.K. Li, ``Chemical Reaction Optimization for Cognitive Radio Spectrum Allocation," in \textit{Proc. IEEE Global Commun. Conf.}, Dec. 2010, pp. 1-5.

\bibitem{AYSLam2011} A. Y.S. Lam, V. O.K. Li, and J. J.Q. Yu, ``Real-Coded Chemical Reaction Optimization," \textit{IEEE Trans. Evol. Comput.}, accepted for publication.

\bibitem{CityBus2011} Citybus Limited and New World First Bus Services Limited, ``Route List", \textit{Available: http://www.nwstbus.com.hk/routes/routesearch.aspx}, access date: Nov. 17$^{th}$, 2011.

\bibitem{JJQYu2011} J. J.Q. Yu, A. Y.S. Lam, and V. O.K. Li, ``Evolutionary Artificial Neural Network Based on Chemical Reaction Optimization," in \textit{Proc. IEEE Congr. Evol. Comp.}, Jun. 2011, pp. 2083-2090.

\bibitem{CJ2012}Code Project, \textit{AI - Simple Genetic Algorithm (GA) to solve a card problem} [Online]. Available: http://www.codeproject.com/KB/recipes/Genetic\_Algorithm.aspx
\end{thebibliography}
\end{document}